\documentclass[11pt, a4paper]{article}
\usepackage[margin=1in]{geometry}
\usepackage{amsmath, amssymb, amsthm}
\usepackage{graphicx}
\PassOptionsToPackage{hyphens}{url}
\usepackage{hyperref}
\usepackage{booktabs}
\usepackage{listings}
\usepackage{xcolor}
\usepackage{caption}
\usepackage{float}
\usepackage{microtype} 
\usepackage{natbib}

\definecolor{codegray}{rgb}{0.5,0.5,0.5}
\definecolor{codepurple}{rgb}{0.58,0,0.82}
\definecolor{codegreen}{rgb}{0,0.6,0}
\definecolor{backcolour}{rgb}{0.05,0.05,0.05}  

\begin{document}

\title{Task-Phase-Conditioned Expert Routing Is Not a Reliable Proxy for Quantization Sensitivity in MoE FFN Tensors}
\author{Yifan Liu \\ \small UIUC MCS Student \small yifanl25@illinois.edu }
\date{\today}
\maketitle

\begin{abstract}
We investigate whether per-layer, per-task-phase (prompt/reasoning/response) expert-routing statistics in a Mixture-of-Experts (MoE) language model can guide layer-wise mixed-precision quantization of routed-expert FFN tensors (down/gate and up). To attain layerwise routing statistics, we build a routing-capture and profiling pipeline for Qwen-3.6-35B-A3B in ik\_llama.cpp~\citep{ikllamacpp}, derive a composite sensitivity score from routing entropy/decisiveness, coefficient of variation (CV), Jaccard/phase-fragility, and top-1 expert strength, and use coordinate descent to tune a layer allocation that respects a fixed average bits-per-weight budget. Notably, the profiling pipeline is applicable to standard top-k MoE architectures. Across independent experimental conditions ($\sim$4.0 bpw flat, $\sim$2.7 bpw steep, and $\sim$2.7 bpw random allocation as a control), reallocation fails to outperform the uniform baseline within each quality tier.
 In the flat regime, reallocation is statistically neutral ($z \approx 0.9$). For the steep tier, the random reallocation performs within 1 SE of the routing-guided reallocation at the same tier, indicating a structural loss at this quality rung caused by mandatory downgrades and Kullback-Leibler Divergence (KLD) tail injection. This aggressive-budget condition shows a decisive loss ($z > 4$ SE) replicated across three independent weight vectors. We argue that the deeper issue is that routing-behavior statistics are an unvalidated proxy for true per-layer quantization cost. 
\end{abstract}

\section{Introduction}

As MoE models have become ubiquitous in deployment, there exists a need for the development of targeted quantization strategies to achieve better KLD and Perplexity (PPL) values. A notable trend has been models becoming increasingly large, where users frequently turn to aggressive quantizations to run these models locally. 

We construct a test harness to evaluate whether MoE routing behavior through different experts for token types and phases can guide budget-neutral reallocation of precision per layer. Specifically, we characterize these phases as prompt, Chain of Thought (CoT) reasoning, and response. We employ the high-quality IQ quantization types present in ik\_llama.cpp~\citep{ikllamacpp}. A natural hypothesis arising from MoE routing is that there exist MoE layers with more distinctive phase-dependent or concentrated routing behavior and that these layers should receive more quantization precision while less distinctive layers should receive more aggressive quantization; we test whether this hypothesis holds. We initially allocated extra precision to the distinctive/fragile layers; however, our best-measured point for KLD actually resulted from a flip in allocation strategy placing more precision on the uniform/robust layers. That direction flip still failed to beat the uniform baseline.

Previous work on MoE quantization has mainly focused on activation-frequency or expert-importance mixed-precision MoEs \citep{mxmoe, gemq, mopeq}, classical rate-distortion (water-filling) bit allocation, and sensitivity-based mixed precision in dense LLMs with direct measures, such as the Hessian and single-layer ablation. These methods condition on either a behavioral activation or importance signal that is not phase-aware (MxMoE, GEMQ) or on a direct sensitivity measurement (MoPEQ and the dense-model methods); none tests whether a phase-conditioned routing proxy can substitute for a direct measurement. 

We do not describe a method that beats existing quantization types and methods. We lay out the precise coordinate descent experiments to find coefficients for our allocation function that tests the hypothesis and analyze the likely causes for failure. During the evaluation of the quantized candidates devised from the reallocation strategy against a corpus of coding text, there were several positions with dramatically large KLD; the 99th-percentile KLD rose from 0.44 (uniform) to $\sim$0.49 on the reallocated candidates.
 Since mean-KLD is tail-dominated the injection of these values increased KLD and PPL against the uniform baseline and decreased top-1 token agreement. The magnitude of these effects varied across quant tier. In the flat regime, the difference in KLD with the uniform baseline was within noise (a neutral result for KLD $z \approx 0.9$), whereas the steep regime shows a decisive loss across all these metrics (KLD $z > 4$ SE).

The rest of this paper will explore the data used, the profiling pipeline, the composite score, and the coordinate-descent search in Section~\ref{sec:methods}. We report the two budget regimes and the random-selection control in Section~\ref{sec:results}. Finally, we provide an analysis of why budget-neutral reallocation loses and its implications for routing statistics as a quantization signal in Section~\ref{sec:discussion}.

\section{Related Work}

\textbf{MoE mixed-precision quantization}. Mixed-precision quantization has become the standard for post-training quantization (PTQ) of MoE LLMs. Existing approaches differ in the signals for bit-width allocation. MxMoE~\citep{mxmoe} combines two strategies, including linear-block sensitivity estimates and a systems approach that leverages expert activation frequency for hardware co-designed allocation, tailoring to the computational capabilities of the deployment environment. GEMQ~\citep{gemq} focuses on global, expert-level importance ranking in a progressive refinement loop to achieve extremely low-bit quantizations, arguing that per-layer local importance estimation does not optimally account for routing, which is perturbed by quantization. MoPEQ~\citep{mopeq} uses Hessian trace approximation to allocate varying bit-widths for experts, noting that expert activation frequency does not predict sensitivity. EAQuant~\citep{eaquant} addresses the problem of activation outliers by employing expert-aware activation smoothing and router-logit alignment, while also focusing on improving calibration data for MoE LLM PTQ. MoEQuant~\citep{moequant} also leverages calibration sampling while contributing affinity-guided quantization to quantify the interplay between individual samples and experts. Both EAQuant and MoEQuant do not propose a sensitivity-based allocation scheme. None of these methods conditions its sensitivity or importance signal on task phase (prompt, reasoning, and response). Building on MoPEQ's finding that expert activation frequency does not predict sensitivity, our approach instead tests whether a phase-conditioned routing-behavior signal can serve as a lightweight proxy for the direct measurements (Hessian trace, per-expert ablation) that MoPEQ and dense-model methods employ.

\textbf{Classical rate-distortion allocation}. Optimal bit allocation given a total-rate constraint is a classical problem in rate-distortion theory~\citep{cover2006elements}. Notably, the water-filling solution~\citep{shoham1988efficient} allocates marginal bits to equalize marginal distortion-reduction per bit across all dimensions when given a per-dimension distortion-rate function. The allocation problem for distributing a fixed average-bpw budget across layers is structurally the same as water-filling; however, the routing-behavior signals (decisiveness, CV, phase-fragility, and top-1 expert strength) we leverage in our composite score are an indirect proxy for the true distortion-rate curve for a layer.

\textbf{Sensitivity-based mixed precision in dense LLMs}. In dense-model quantization, Hessian-trace and eigenvalue approximation methods such as HAWQ~\citep{hawq} and GPTQ's~\citep{frantar2023gptq} Hessian-guided error compensation, along with single-layer ablation measurements to isolate the contribution of one component to output error serve as the standard methods to directly estimate per-weight or per-layer sensitivity. The accuracy of these direct measurements depends on a backward pass or ablation sweep per layer. Other more ubiquitous methods include imatrix~\citep{ikllamacpp}, which can be applied to both MoE and dense models by calculating per-channel activation statistics through calibration sampling. This paper relies on a purely behavioral routing statistic computed through forward passes alone (somewhat similar to imatrix) to substitute for direct sensitivity measurement in a MoE setting. 

\section{Methods}
\label{sec:methods}

\subsection{Model and Setup}

The model employed throughout all the experiments is Qwen-3.6-35B-A3B~\citep{qwen36model}. This model has 40 layers plus a separate MTP head (blk.40); dynamic allocation applies to blk.0-38, with the terminal layer blk.39 held at baseline. The model is quantized with an imatrix and run in ik\_llama.cpp~\citep{ikllamacpp}, the engine that we modify for our experiments. The imatrix supplies per-channel activation statistics from a calibration set to guide quantization scaling; it is the standard llama.cpp calibration mechanism and is separate from the routing-capture allocation signal developed in Section~\ref{sec:composite}.

\subsection{Data Pipeline and Bug Fixes}

We curate a pool of 186 coding prompts from HumanEval-X~\citep{zheng2023codegeex} (Python, C++, Java, JS, Go), HumanEval-Rust~\citep{humanevalrust}, and LiveCodeBench Lite~\citep{jain2024livecodebench}, spanning easy, medium, and hard difficulty. For the profiling pass through the modified ik\_llama.cpp engine, we stratify-sample 30 prompts from this pool, balanced across source, language, and difficulty.

We patch ik\_llama.cpp to capture the per-token top-k expert selections and gate probabilities at each MoE block, allowing us to characterize the routing behavior of Qwen-3.6-35B-A3B. For efficiency gains, the strategy we use is surrogate routing capture. The model is first quantized to $\sim$4 bpw per tensor with a Q8\_0 precision router and attention/SSM tensors to ensure high fidelity routing with the routed experts quantized to IQ4\_KS (down) and IQ3\_K (gate/up). Since the Q8\_0 router decides the actual top-k experts which fire per token, the signal is still faithful when the routed experts are quantized. The objective of surrogate routing capture was to ensure that it was possible to capture routing data on systems where running the full quality BF16 is not possible. 

We resolve several bugs regarding our patch. Initially, we encountered issues with identifying the boundaries of prompts in the captured data. To be able to separate each prompt from the routing capture data, we added per-request IDs for each prompt profiled. One bug caused routing capture data to be dropped without error. We discovered that the fix was to turn graph reuse off. This change decreases decode performance to ensure proper routing capture data. Another separate bug also caused silent failures for the pipeline to capture prompts of certain lengths. The source of this bug was when prompt token length exceeded the value set for ubatch-size which corrupted the capture. It was essential to set both the batch and ubatch size to be longer than the maximum token count of the 30 coding prompts. 

\begin{figure}[ht]
\centering
\includegraphics[width=\textwidth]{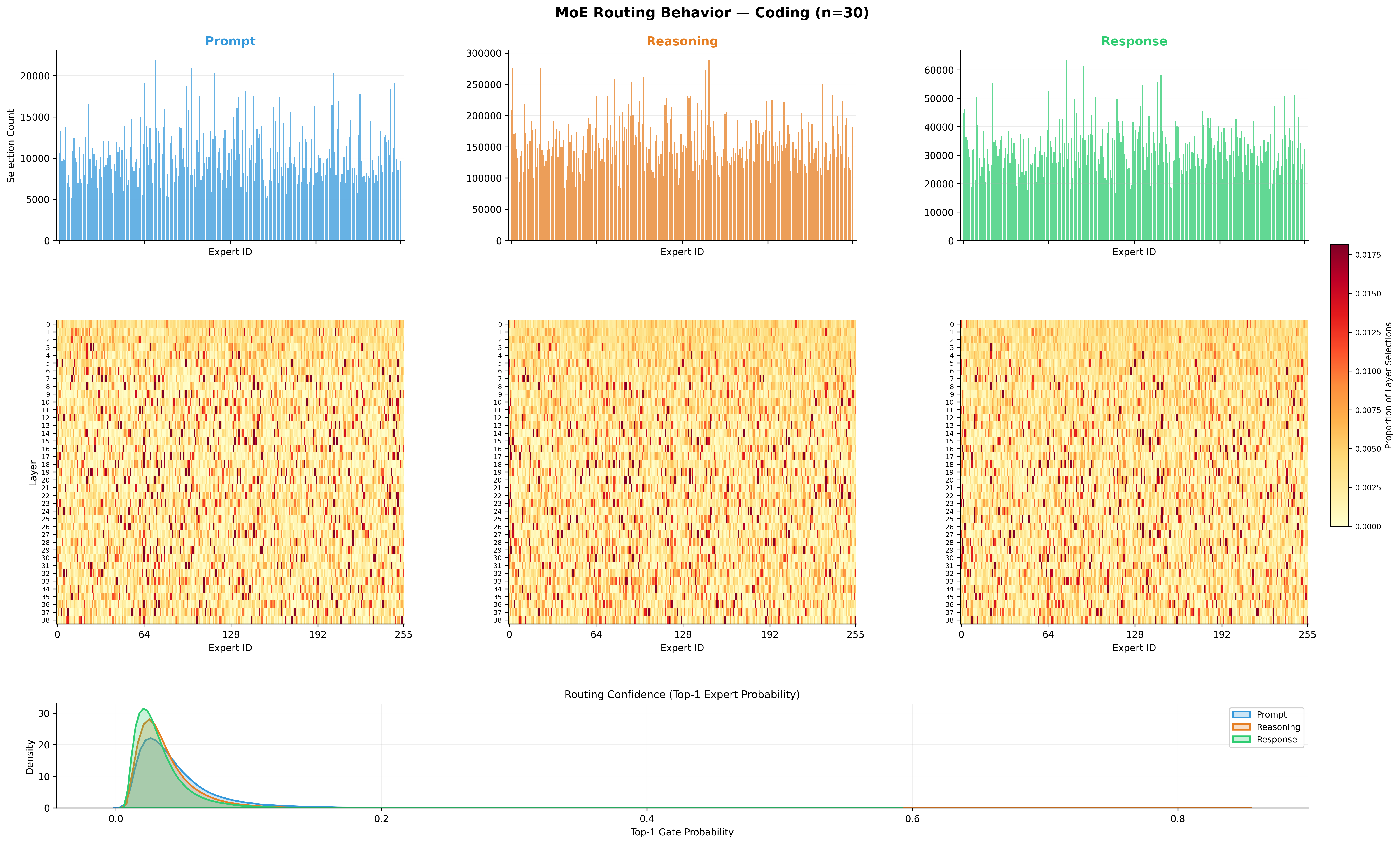}
\caption{Per-expert routing behavior across the three task phases. Top row shows aggregate expert selection counts per phase. Middle row shows per-layer expert-usage heatmaps, one row per profiled layer. Bottom row shows KDEs of top-1 gate probability. Data spans 30 coding prompts, 39 profiled layers, and 256 routed experts. Prompt-phase routing is visibly less diffuse than the reasoning and response phases.}
\label{fig:routing}
\end{figure}

During the collection phase, we obtain expert coverage percentage, top-1 fraction, normalized entropy, and cross-prompt Jaccard (mean/std) and cross-phase pairwise Jaccard. These scores are combined into four signals to describe per-layer and per-phase statistics. These signals include entropy (decisiveness), CV (concentration), cross-phase Jaccard (phase-fragility), and top-1 gate prob (confidence). These become the four scores which we assign coefficients and optimize for in coordinate descent. In \autoref{tab:segment_stats} and \autoref{fig:routing}, we describe a high-level summary of the results from the routing capture pipeline. The prompt-phase coverage floor shows that some experts never fire in prompt-processing across all thirty prompts. Also, there is a position-count disparity across the different phases. 

\begin{table}[ht]  
\centering 
\begin{tabular}{lccccc}
\toprule       
segment & $n_{\text{positions}}$ & med.\ top-1 frac. & med.\ $H_{\text{norm}}$ & expert coverage \\ 
\midrule
prompt    & 8{,}508   & 0.0558 & 0.863 & 0.89--1.00 \\   
reasoning & 128{,}287 & 0.0322 & 0.907 & 0.996--1.00 \\  
response  & 26{,}721  & 0.0366 & 0.912 & 0.996--1.00 \\       
\bottomrule 
\end{tabular}               
\caption{Per-phase routing statistics over 30 coding prompts (39 profiled layers each).   Prompt-phase routing is markedly less diffuse (lower $H_{\text{norm}}$) and leaves some experts never selected (coverage down to 0.89), unlike the reasoning/response phases.}
\label{tab:segment_stats}  
\end{table}  

\subsection{Segmentation}
The profiling is segmented into prompt, reasoning, and response through two boundary rules. The prompt is designated as the prefill region and defined by the request's prompt\_tokens length, whereas the reasoning is the span from the end of the prompt up to and including the first $<$/think$>$ token. The remainder of the generation is the response. If any generation lacks a $<$/think$>$ token, then all post-prompt content is folded into the response; in practice, all 30 prompts emit a close delimiter. We constrained the model's server-side reasoning budget to 8192 tokens. The maximum total tokens per request was set to 32768, ensuring that the model finished its reasoning and emitted a full response. Since each segment differs substantially in token count (8.5k / 128k / 26.7k), per-layer statistics were computed and z-scored within each segment.

\subsection{Composite Sensitivity Score}
\label{sec:composite}

To allocate bits and different quantizations to routed-experts tensors (down, and gate/up), we devise a composite score function. This score consists of four signals each representing a property of a layer's expert-usage distribution, including decisiveness (sharp routing), concentration, phase-fragility (how differently the model routes across phases), and confidence. 

With each segment having a different aggregate token count (differing up to $\sim$15x) each signal needs to be z-scored for normalization within its own scale. Since we take the maximum composite score (max-aggregation) across all three segments, the z-scoring prevents any composite score from dominating through volume of tokens processed, facilitating fair comparison across the different segments.

The weighted composite function is shown in Equation~\eqref{eq:composite}. The weights allow each signal's relative importance to be tuned. Iterative coordinate descent is applied to determine the optimal weights.

\begin{equation}
  \begin{aligned}
    d_l \;=\; 1 - H_l, \qquad c_l \;=\; \mathrm{CV}_l, \qquad  
    f_l \;=\; 1 - \min\!\big(J_l^{rr},\, J_l^{pr}\big), \qquad t_l \;=\; \tau_l \\[4pt]
    z^{(s)}_l(x) \;=\; \dfrac{x_l - \bar{x}^{(s)}}{\sigma^{(s)} + \epsilon} \\[4pt]
    S^{(s)}_l \;=\; w_d\, z^{(s)}_l(d_l) + w_c\, z^{(s)}_l(c_l) + w_p\, z^{(s)}_l(f_l) + w_t\, z^{(s)}_l(t_l) \\[4pt]
  \end{aligned}
  \label{eq:composite}
\end{equation}

where $H_l$ is the normalized routing entropy, $\mathrm{CV}_l$ the coefficient of variation of expert usage, $J_l^{rr}$ and $J_l^{pr}$ the cross-phase pairwise Jaccard overlaps (reasoning $\cap$ response, prompt $\cap$ reasoning), $\tau_l$ the top-1 expert fraction, $x \in \{d, c, f, t\}$ a signal (so $x_l$ is its value at layer $l$), $\bar{x}^{(s)}/\sigma^{(s)}$ the mean/std of signal $x$ within segment $s$, and $\epsilon$ a small constant.

The purpose of the max-aggregation is to identify if a layer is sensitive or extreme in any phase according to the four signals. Whether a high score receives more or fewer bits is a free allocation-direction choice, handled in Section~\ref{sec:allocation}. 

Because the downstream allocation in Section~\ref{sec:allocation} consumes only $\mathrm{rank}(S_l)$, the ordering of layers by score and not values, the composite score is invariant to any affine rescaling of its inputs. We exploit this characteristic to fix $w_d \equiv 1$, so the composite score is parameterized by a four-component weight vector $\mathbf{w} = (w_d, w_c, w_p, w_t)$ with $w_d$ held at 1; the search is therefore over the three free ratios $\{w_c, w_p, w_t\}$ only. The 3-axis coordinate descent in Section~\ref{sec:coord} is consequently well-posed.

\subsection{Allocation Mechanism}
\label{sec:allocation}

For every profiled layer, we choose a down tier and gu tier pair. Specifically, we bundle the gate and up tensors together and quantize them at the same tier. The byte model is $down + 2\cdot gu$ per layer. The multiplier 2 is deliberate since gate and up are separate tensors.  

To designate different quantization types to allocate to the layers, we draw from ik\_llama.cpp's IQK family; the per-tensor bpw of every tier used in this work is listed in Appendix A (Table~\ref{tab:iqk-bpw}). Each layer's allocated tier is within a small band ($\pm$1-2 rungs $\approx$ 0.25-0.5 bpw per rung) of that regime's baseline so the reallocation is a local rearrangement. We analyze the effects of dynamic layer-wise quantization at two quality regimes.  

\begin{table}[ht]
\centering
\footnotesize
\begin{tabular}{@{}l p{6.0cm} l l c@{}}
\toprule
Regime & Ladder (rungs) & Down base & GU base & Combined bpw/layer \\
\midrule
Flat ($\sim$4 bpw)   & \texttt{iq3\_ks$\cdot$iq3\_k$\cdot$iq4\_kss$\cdot$iq4\_ks$\cdot$iq4\_k} & \texttt{iq4\_ks} & \texttt{iq3\_k}  & 11.13 \\
Steep ($\sim$2.7 bpw) & \texttt{iq2\_ks$\cdot$iq2\_k$\cdot$iq2\_kl$\cdot$iq3\_ks$\cdot$iq3\_k} & \texttt{iq3\_ks} & \texttt{iq2\_kl} & 8.57 \\
\bottomrule
\end{tabular}
\caption{Quantization regimes and corresponding ladder configurations. \texttt{gu} = the bundled gate/up tier.}
\label{tab:quantization-regimes}
\end{table}

We run a budget-fit algorithm that consists of the following steps. The layers are first ranked by their composite-score percentile. The layers are then split into a fragile pool of the low-ranked layers and a robust pool of the high-ranked ones. We downgrade the layers of the fragile pool to save bytes and build a byte bank surplus. We then use this surplus to fund the upgrades of the robust pool, one rung at a time, while keeping the total routed-expert bytes at or below the budget fixed by the uniform baseline, so every candidate is budget-neutral by construction. Several quality parameters such as max\_step (max rungs a layer may move) and stay\_frac (fraction of layers that may move at all) allow us to vary experiments for aggressive and conservative reallocation.

The importance of budget-neutral quantizations cannot be understated, as it forms the basis for comparable and fair quality (KLD and PPL) comparisons between the quantizations generated from the budget-fit algorithm and the uniform baseline. We define budget-neutral as size-matched to the uniform baseline of that quality regime with approximately the same amount of bits just rearranged. This definition is enforced in the algorithm with the two gates, including a pre-quantize dry-run projection check for the final size of the model post quantization and a post-quantize measured-byte check. Both of these checks are compared against the uniform baseline's bytes. 

Another parameter we set is whether the extra bits go to the fragile layers or the robust layers. We ran experiments in both directions; the favorable direction is discussed in Section~\ref{sec:results}. To avoid re-benchmarking identical allocations, we canonicalize each tier\_map to a sorted JSON form, hash it with SHA-1, and use the resulting hash as the cache key. If a tier\_map is already in the cache, its metrics are reused with no re-quantization, while a miss triggers a single quantize-and-evaluate. Two distinct weight vectors that yield the same tier\_map therefore collapse to one entry so redundant allocations are never re-quantized.

\subsection{Coordinate Descent Search}
\label{sec:coord}

The coordinate descent process optimizes the 4 weights in the composite score; however, we only tune the 3 free weight ratios from Section~\ref{sec:composite} with $w_d$ anchored to 1. We tune each free ratio with coordinate descent cyclically, one axis at a time, where the objective is minimum KLD. The top-1 token agreement and PPL are also recorded but not used to dictate descent direction. Each weight vector maps to a tier\_map, as mentioned in Section~\ref{sec:allocation}, and each tier\_map is evaluated one time to obtain the KLD. 

A different grid of multipliers is implemented for each round. In the first round, which is coarse, each axis tests multipliers $\{0.5, 1.5, 3.0\}$ of its value. In the second round, which is adaptive, the grid of an axis is chosen by the winner of the first round. An interior winner is assigned a grid of $\{0.75, 1.0, 1.25\}$. A high-edge winner gets a grid of $\{1.0, 1.5\}$ which extends the search up, capped at 4.5x. A low-edge winner gets a grid of $\{1.0, 0.5, 0.25\}$, extending the search down. In the worst case, the coordinate search explores at most 18 distinct quantizations (3 axes x 3 candidates x 2 rounds). The number of quantizations is usually fewer after the caching strategy in Section~\ref{sec:allocation} and early-stop is applied. The early-stop strategy defines a stopping condition as a full pass that changes nothing, where there is no KLD improvement. As a control experiment, we use a seed-frozen, depth and budget-matched random-layer-selection variant evaluated through the same pipeline as a control. 

Our coordinate descent implementation is resumable and cache-aware, as intermediate states are written to temporary JSON tracking files as checkpoints. This design allows for pausing and resuming the GPU-driven evals. Since the objective is a plateau, many weight vectors map to the same tier\_map and hence the same KLD. The search is also dedupe-aware, skipping any candidate whose tier\_map is already measured. An example of the full logs can be found in Appendix B.

\subsection{Evaluation Methodology}

To evaluate each quantization generated by the budget-fit algorithm, we first construct the high-precision KLD reference by running the model quantized uniformly to Q8\_0 over a coding corpus. We use Q8\_0 as a proxy for BF16 quality; prior work on Qwen3 quantization reports near-lossless performance at moderate-to-high bit-widths including 8-bit precision, with degradation emerging primarily under ultra-low-bit settings~\citep{zheng2025qwen3quant}, and independent KL-divergence measurements on Qwen-3.6 similarly find that 8-bit quantization tracks BF16 closely, with residual divergence roughly an order of magnitude smaller than 4-bit quantization~\citep{mcleod2026kld}.

We measure each candidate with a single llama-perplexity run using \texttt{--kl-divergence} and \texttt{--kl-divergence-base} against the Q8\_0 reference logits. This single run yields mean KLD, PPL, and top-1 agreement, along with the KLD tail: the 99th and 95th percentiles, the median, and the maximum. The llama-perplexity flags are shared across the Q8\_0 reference, the candidates, and the uniform baseline through a single function, so all three are computed with identical context, layer placement, and batch settings to ensure direct comparability. 

The evaluation corpus is drawn from The Stack v3 by streaming the dataset from HuggingFaceCode~\citep{stackv3}. It is held out from the profiling data, which originates from HumanEval-X and LiveCodeBench, so no prompt in the evaluation set was seen during routing capture. To reduce training-data contamination we exclude repositories created before June 2024, restrict to permissively-licensed code, and drop vendored third-party files. To promote representational diversity, a per-language token budget balances coverage across 21 languages, including niche languages, while seeded shuffling and multi-shard sampling provide additional diversity during data selection. The final corpus is $\sim$331K tokens ($\sim$647 chunks at context 512), comparable in size to the widely-used wikitext-2 test set~\citep{merity2016pointer}.

\subsection{A Methodological Note on Validation-Set Bias}

The coordinate-descent search is a plateau objective over a small weight space. Initially, we run the evaluations on a frozen 12K-token slice ($\sim$24 chunks at context 512) to keep GPU evaluation cheap. We unfortunately note that the slice does not serve as an accurate proxy for the true full-corpus KLD. Specifically, the uniform baseline's KLD at the IQ3\_K quality regime was 0.019325 on the 12K slice but 0.014725 on the full 331K-token corpus. The full-corpus KLD is $\sim$24\% lower than the slice estimate. Since the relative ranking of candidates against the uniform baseline also did not transfer from the slice to the full corpus, we report all final results on the full corpus and acknowledge that the slice is only used to drive the descent.

\section{Results}
\label{sec:results}

\subsection{Flat Regime (\texorpdfstring{$\sim$}{~}4 bpw, Compressed Ladder)}

Quantizations in this regime along with the uniform baseline are at the flat region of the Pareto curve. Across all 8 quantized candidates benchmarked, in Table~\ref{tab:flat}, we observed that none of them beat the uniform baseline that had down experts at IQ4\_KS and gate/up experts at IQ3\_K. The best candidate (C1) achieves a result close to neutral (+0.94 SE) on KLD compared to the uniform baseline. At $\sim$4 bpw, every layer is already near its per-layer optimum, so moving bits between layers trades a small gain on a few layers against a small loss on the rest. This reallocation nets to roughly zero change in KLD.  

\begin{table}[ht]
\centering
\footnotesize
\begin{tabular}{@{}lcc@{}}
\toprule
Allocation & KLD ($\pm$ SE) & Top-1 (\%) \\
\midrule
Uniform (baseline) & 0.014725 $\pm$ 0.000430 & 97.5 \\
\midrule
C1 & 0.015276 $\pm$ 0.000400 & 97.4 \\
C2 & 0.015327 $\pm$ 0.000405 & 97.3 \\
C3 & 0.015362 $\pm$ 0.000407 & 97.3 \\
C4 & 0.015363 $\pm$ 0.000403 & 97.4 \\
C5 & 0.015507 $\pm$ 0.000423 & 97.3 \\
C6 & 0.015662 $\pm$ 0.000459 & 97.3 \\
C7 & 0.015709 $\pm$ 0.000451 & 97.3 \\
C8 & 0.015851 $\pm$ 0.000457 & 97.3 \\
\bottomrule
\end{tabular}
\caption{Flat regime ($\sim$4 bpw, compressed ladder). Uniform baseline vs. the 8 coordinate-descent allocations C1--C8 ordered by KLD, on the full 331K-token corpus. No allocation beats the uniform baseline. The best candidate is statistically neutral. PPL is a wash across all candidates at $\sim$2.00 and is omitted.}
\label{tab:flat}
\end{table}

\subsection{Steep Regime (\texorpdfstring{$\sim$}{~}2.7 bpw, Low-Bit)}
\label{sec:steep}

Quantizations in this regime along with the low-bit uniform baseline are on the steep region of the Pareto curve (past the elbow). Notably, in this quality tier, the loss is decisive ($z \approx 4.3$ SE, combined). As shown in Table~\ref{tab:steep}, the loss is tail dominated from the floor tier quality at IQ2, and the 99th percentile KLD shows substantial increases, rising from 0.437 (uniform) to 0.487--0.492, while the median KLD stays at most 0.0019. The tail dominates because the mean of 0.034 is $\sim$20$\times$ the median. We analyze the exact mechanism for the mean KLD degradation in this regime in Section~\ref{sec:discussion}. The top-1 token agreement percentage also drops modestly from 95.8\% to 95.6\%, showing the effects of quantization noise at the token level. 

\begin{table}[ht]
\centering
\footnotesize
\begin{tabular}{@{}lccc@{}}
\toprule
Allocation & KLD ($\pm$ SE) & KLD p99 & Top-1 (\%) \\
\midrule
Uniform U (baseline) & 0.034321 $\pm$ 0.000608 & 0.437 & 95.8 \\
\midrule
cv 0.5 & 0.038249 $\pm$ 0.000659 & 0.490 & 95.6 \\
cv 1.5 & 0.038142 $\pm$ 0.000665 & 0.487 & 95.6 \\
cv 3.0 & 0.038223 $\pm$ 0.000662 & 0.492 & 95.6 \\
\bottomrule
\end{tabular}
\caption{Steep regime ($\sim$2.7 bpw, low-bit, IQ2 floor tier). Uniform baseline U vs. the 3 score-guided allocations on the full corpus. All candidates lose decisively. The KLD tail (p99) rises on every candidate.}
\label{tab:steep}
\end{table}

\subsection{Random-Selection Control}

To determine if our method held any statistical significance, we compare our score guided approach against two quantization recipes generated with random seeds. Holding depth, tiers, and budget fixed and randomizing the assignment of quantization types to the layers, the random candidates land within $\sim$1 SE of (and marginally below) the score-guided ones. This result informs us that the reallocation structure is responsible for the loss, where the mandatory IQ2 downgrades that were required to fund the upgraded-quality layers inject KLD-tail positions. Decisively, the composite score adds no selection signal over random layer selection. Table~\ref{tab:control} reports these random candidates alongside the score-guided set and the uniform baseline.

\begin{table}[ht]
\centering
\footnotesize
\begin{tabular}{@{}lccc@{}}
\toprule
Allocation & KLD ($\pm$ SE) & KLD p99 & Top-1 (\%) \\
\midrule
Uniform U (baseline) & 0.034321 $\pm$ 0.000608 & 0.437 & 95.8 \\
\midrule
random, seed 42 & 0.037728 $\pm$ 0.000642 & 0.483 & 95.7 \\
random, seed 1337 & 0.037808 $\pm$ 0.000617 & 0.495 & 95.6 \\
\midrule
\textit{ref: score-guided (cv 0.5--3.0)} & 0.03814--0.03825 & 0.487--0.492 & 95.6 \\
\bottomrule
\end{tabular}
\caption{Random-selection control ($\sim$2.7 bpw). Depth, tiers, and budget are matched to the score-guided candidates. Only layer selection is randomized across 2 seeds. Random lands within $\sim$1 SE of the score-guided candidates, showing the composite score carries no selection signal over random.}
\label{tab:control}
\end{table}

\section{Discussion}
\label{sec:discussion}

\subsection{Why Budget-Neutral Reallocation Loses}
\label{sec:budget}
Fundamentally, budget-neutral reallocation loses because of its zero-sum structure: every upgrade is funded by downgrades on other layers. In the steep regime, for example, upgrading 17 layers by one rung ($+1.25$ bpw per layer) required downgrading 19 layers to the IQ2 floor tier ($-1.12$ bpw per layer), with 3 layers staying at baseline; the gains and costs cancel almost exactly, so every candidate lands at the uniform budget. This finding leads to the fact that the net $\Delta \mathrm{KLD} = \sum_l g_l - \sum_l c_l$, where $g_l$ and $c_l$ are the per-layer marginal gains and costs. To produce a win against the uniform baseline, the reallocation criterion would need to rank layers by their true marginal KLD-per-byte. 

Water-filling theory states that optimal allocation equalizes marginal distortion-reduction per bit. Our proxy composite score utilizes routing behavior measures (entropy, CV, phase-fragility, top-1 strength) which are not a direct measurement of quantization distortion. The premise of our experiments was based on the hypothesis that layerwise phase-distinctive routing had a degree of correlation with the cost of reducing or increasing the bpw for a specific layer. There was never a theoretical basis for this expectation, and our empirical results showed it to be unsupported as well.  

In the flat regime, the effect of the net $\Delta$KLD can be seen directly. At this quality tier all layers are near optimum on the uniform baseline. Therefore, any marginal gains from our allocation are immediately nullified by the marginal costs. The overall KLD nets to $\sim$ zero, with all 8 candidates actually scoring slightly worse ($z = +0.94$ to $+1.79$). In the steep regime, there was a convex cost curve and the IQ2 floor cliff. This characteristic led to downgrades costing disproportionately more than upgrades gain; therefore, any nontrivial reallocation loses. 

Naturally, we questioned the entire premise with the random-selection control. If the routing statistics tracked quantization sensitivity even weakly, the score-guided allocations would have beaten random selection at matched depth and budget. Since random selection lands within $\sim$1 SE of (and marginally better than) the score-guided candidates, at 0.03773/0.03781 KLD versus 0.03814--0.03825, we acknowledge that the composite score adds no selection signal over random. 

\subsection{Relationship to Floor-Tier Cliff}
As we discussed in Section~\ref{sec:budget}, there are two compounding effects: the per-tensor IQ2 floor quality cliff, and the proxy failing to identify differentially-sensitive layers.  

The floor-tier cliff can be quantified by its tail KLD injections. With 19 out of 39 profiled layers requiring mandatory downgrades, the 99th percentile KLD rises from 0.437 to 0.483--0.495 on every reallocated candidate, score-guided and random alike. The baseline KLD distribution is already heavy-tailed: median 0.0016, 95th percentile 0.119, maximum 26.4, hence the $\sim$20$\times$ mean/median gap reported in Section~\ref{sec:steep}. The mandatory downgrades act on this tail rather than on the bulk. The median moves only from 0.0016 to at most 0.0019, and the maximum stays within 26.4--27.5, so neither the typical position nor the single worst case is what degrades. Instead, the distribution develops greater mass in the tail, with more positions entering the catastrophic regime, and this shift accounts for the mean-KLD loss of 0.0034--0.0039.

The random-selection allocations lose to the uniform baseline because the cliff toll is incurred per tensor rather than per layer. At this depth, any budget-neutral reallocation must therefore downgrade 19 layers to the floor tier, regardless of which layers are selected. Swapping which layers receive the floor-tier IQ2 quality does not change the cost, because the tail injection is a property of the tier itself rather than of the specific layers that carry it. The implication is that below the elbow of the Pareto curve, reallocation within a fixed budget does not help. The only mitigation of catastrophic tail injection is to raise the floor globally, providing more budget.

\subsection{Limitations}

A primary limitation of our study was the application of the composite score budget-reallocation strategy to a single model, Qwen-3.6-35B-A3B, which is a hybrid linear-attention MoE. We do not claim that our results can generalize to other models or scales. A cross-model transfer to the much larger GLM-5.3~\citep{glm53model}, a MoE language model with 753B total and 40B active parameters, was planned but not executed.

Since our per-layer, phase-based routing profiling ran on a surrogate quantization (down experts at IQ4\_KS, gate/up at IQ3\_K, with a Q8\_0 router), it is possible that the routing was itself perturbed by quantization. There was no comparison performed to check the routing agreement versus a full-precision BF16 version of the model.

There was also no direct-sensitivity ground truth. We disclose the limitation of having no single-layer ablation study and the lack of Hessian correlation values. We are unable to bound what a direct quantization-sensitivity signal would have achieved; we can only conclude that this proxy achieves nothing. 

For the data, we constructed additional datasets for multiple topics (reasoning, roleplay, chat, and trivia); however, the experiments in this paper only utilized the code task domain for both profiling and evaluation. The number of samples profiled was 30, and no tests were done to validate whether this was the optimal number of profiling samples. During the layerwise quality allocations, we opted to hold the terminal layer (blk.39) at the same quality as the uniform baseline; therefore, 39 of the 40 layers were profiled and received allocations. It is possible that including the terminal layer in the allocation could have influenced the results slightly. Segmentation also depended on the $<$/think$>$ delimiter to split generations into prompt, reasoning, and response, which all 30 profiling prompts happened to emit; a segmentation using only the prompt and response segments (e.g., for models without a reasoning trace) could have produced different sensitivity signals for the composite score. 

\section{Conclusion}

Task-phase-conditioned routing statistics, despite a carefully instrumented pipeline spanning routing capture, composite scoring, and budget-neutral coordinate descent, do not translate into a useful equal-budget allocation criterion for layer-wise mixed-precision quantization of the routed-expert FFN tensors. The score-guided allocations are neutral at $\sim$4 bpw and decisively harmful at $\sim$2.7 bpw. We underscore that the mechanism behind these results is structural: mandatory downgrades cause the KLD tail injection. The random-selection control shows that the heuristic is not at fault. We report these results to show that the intuitive connection presumed to exist between routing behavior and layer-wise quantization sensitivity does not hold. Future work would need to explore the connection at a higher level of granularity, potentially allocating precision to individual experts within a layer, and to validate candidate signals against directly measured layer sensitivity. 

\sloppy
\bibliographystyle{plainnat}
\bibliography{refs.bib}

\appendix

\section{Full Launch Configuration}
\label{app:launch}

\subsection*{IQK tier bandwidths used}

\begin{table}[ht]
\centering
\footnotesize
\begin{tabular}{@{}lll@{}}
\toprule
Tier & bpw/tensor & Used in \\
\midrule
\texttt{IQ2\_KS}  & 2.1875 & steep ladder (floor) \\
\texttt{IQ2\_K}   & 2.375  & steep ladder \\
\texttt{IQ2\_KL}  & 2.69   & steep ladder (gu base) \\
\texttt{IQ3\_KS}  & 3.19   & both ladders (steep down base) \\
\texttt{IQ3\_K}   & 3.44   & both ladders (flat gu base) \\
\texttt{IQ4\_KSS} & 4.0    & flat ladder \\
\texttt{IQ4\_KS}  & 4.25   & flat ladder (down base) \\
\texttt{IQ4\_K}   & 4.5    & flat ladder \\
\bottomrule
\end{tabular}
\caption{Per-tensor bits-per-weight of the IQK quantization tiers used in this work, as reported by ik\_llama.cpp. All other tensors (attention, SSM, shared experts, embeddings, MTP head) are held at fixed tiers (Q8\_0 or IQ6\_K) outside the reallocation budget.}
\label{tab:iqk-bpw}
\end{table}

Each candidate quantization is produced by a single \texttt{llama-quantize} invocation with a per-layer \texttt{--custom-q} string and the shared imatrix. The perplexity/KLD measurement is a single \texttt{llama-perplexity} run against the Q8\_0 reference log-probs. All flags below are identical across the reference, every candidate, and the uniform baseline (built by the shared \texttt{\_ppl\_common} function in \texttt{driver.py}) to ensure direct comparability.

\subsection*{Quantization}

\begin{lstlisting}[language=bash]
llama-quantize \
    --custom-q "$CUSTOM_Q" \
    --imatrix qwen36moe.imatrix \
    qwen3.6-35B-A3B-bf16.gguf \
    cand_<hash>.gguf \
    IQ3_K 128
\end{lstlisting}

The \texttt{--custom-q} string is a comma-separated list of regex-weight=tier assignments. Routed-expert tensors (\texttt{ffn\_down\_exps} and \texttt{ffn\_(gate|up)\_exps}) receive per-layer tiers chosen by the budget-fit algorithm; all other tensors (attention, SSM, shared experts, embeddings, MTP head) are held at fixed tiers (Q8\_0 or IQ6\_K). A \texttt{--dry-run} size gate is run first, and the candidate is refused if its projected byte size exceeds the uniform baseline.

\subsection*{Perplexity / KLD measurement}

\begin{lstlisting}[language=bash]
llama-perplexity \
    -m cand_<hash>.gguf \
    -f eval_corpus.txt \
    -c 512 -ngl 999 \
    --threads 6 \
    -fa on \
    -b 8192 -ub 8192 \
    -muge -gr -ger \
    --kl-divergence \
    --kl-divergence-base q80\_logits.dat
\end{lstlisting}

The key settings are context 512, all layers on GPU, 6 CPU threads, flash-attention on, batch/ubatch 8192, and the \texttt{-muge -gr -ger} performance flags. The \texttt{--kl-divergence-base} file (\texttt{q80\_logits.dat}) is the log-prob output of the Q8\_0 reference model on the same corpus, saved via \texttt{--save-all-logits}. The single run yields mean KLD, PPL, top-1 agreement, and the KLD tail (p99, p95, median, max).

\section{Coordinate-Descent Log}
\label{app:cdlog}

Table~\ref{tab:cdlog} reproduces the full search trajectory for the flat-regime coordinate descent (6 axis-steps across 2 rounds). The initial weight vector is $\mathbf{w} = (1, 1, 1, 1)$ with $w_d$ anchored at 1. The last column indicates whether the axis winner differed from the current value.

\begin{table}[ht]
\centering
\footnotesize
\begin{tabular}{@{}clccc@{}}
\toprule
Round & Axis & Winning multiplier & KLD & Changed \\
\midrule
1 & $w_c$ (CV)    & $\times 1.5$ & 0.015276 & yes \\
1 & $w_p$ (phase) & $\times 0.5$ & 0.015363 & yes \\
1 & $w_t$ (top-1) & $\times 1.5$ & 0.015327 & yes \\
\midrule
2 & $w_c$ (CV)    & $\times 0.75$ & 0.015276 & yes \\
2 & $w_p$ (phase) & $\times 1.0$  & 0.015276 & no \\
2 & $w_t$ (top-1) & $\times 0.75$ & 0.015276 & yes \\
\bottomrule
\end{tabular}
\caption{Flat-regime coordinate-descent trajectory across 3 free axes and 2 rounds. The objective is full-corpus KLD. The KLD column reports the candidate evaluated at each step. A step whose candidate is worse than the current best (e.g., round-1 $w_p$) is recorded but not accepted. Thus, the accepted KLD never regresses.}

\label{tab:cdlog}
\end{table}

The search converges to a plateau where multiple weight vectors map to the same tier\_map (same SHA-1 hash), hence the same KLD. In this trajectory, round-2 \texttt{cv}$\times0.75$ and \texttt{top1}$\times0.75$ both map to hash \texttt{c5c523c8$\ldots$} (KLD 0.015276).  The best candidate is reported as C1 in Table~\ref{tab:flat}. The early-stop condition, which consists of a full pass with no KLD improvement, is not triggered because the 3 free axes still oscillate among distinct tier\_maps; however, the objective surface is flat, meaning that all 8 final candidates fall within $\sim$0.9\,KLD-SE of each other.

\section{Dedupe Cache Schema}
\label{app:cache}

The dedupe cache maps a canonical allocation hash to its measured metrics. The hash is computed as:

\begin{lstlisting}[language=Python]
canon = json.dumps(
    {str(k): v for k, v in sorted(tier_map.items())},
    sort_keys=True,
)
hash = hashlib.sha1(canon.encode()).hexdigest()
\end{lstlisting}

where \texttt{tier\_map} is a dictionary mapping layer index to a \texttt{(down\_tier, gu\_tier)} string pair. Two distinct weight vectors that yield the same per-layer tier assignment share a single cache entry. Each cache value is a JSON object with fields illustrated in Table~\ref{tab:cacheschema}.

\begin{table}[ht]
\centering
\footnotesize
\begin{tabular}{@{}ll@{}}
\toprule
Field & Description \\
\midrule
\texttt{kld}        & Mean KLD (nats/token), full corpus \\
\texttt{kld\_se}     & Standard error of KLD (across chunks) \\
\texttt{kld\_p99}    & 99th-percentile per-position KLD \\
\texttt{kld\_p95}    & 95th-percentile per-position KLD \\
\texttt{kld\_median} & Median per-position KLD \\
\texttt{kld\_max}    & Maximum per-position KLD \\
\texttt{ppl}        & Perplexity on the evaluation corpus \\
\texttt{top1}       & Top-1 token agreement (\%) vs.~Q8\_0 reference \\
\texttt{bytes}      & Measured GGUF file size (bytes) \\
\texttt{ref\_bytes} & Uniform baseline file size (bytes) \\
\texttt{over\_budget} & Whether \texttt{bytes} exceeds \texttt{ref\_bytes} \\
\bottomrule
\end{tabular}
\caption{Dedupe cache value schema. Fields present in the low-bit cache (\texttt{results\_lowbit.json}). The flat-regime cache (\texttt{results\_full.json}) omits the KLD-tail fields.}
\label{tab:cacheschema}
\end{table}

The cache is a flat JSON file: a single object mapping hashes to metrics. A cache hit (the hash is already present) short-circuits the GPU pipeline entirely; a miss triggers a single quantize-and-evaluate and appends the result. The cache is shared across all experiments and regimes. Thus, re-running the search after interruption or on a different machine never re-quantizes a previously measured allocation.

\end{document}